%% file: main.tex
\documentclass[a4paper,fleqn]{cas-dc}

\usepackage[numbers,sort&compress]{natbib}
\usepackage{amsmath}
\usepackage{amssymb}
\usepackage{booktabs}
\usepackage{colortbl}
\usepackage{graphicx}
\usepackage{multirow}
\usepackage{xcolor}
\usepackage{hyperref}
\usepackage{algorithm}
\usepackage{tabularx}
\usepackage{algorithmicx}
\usepackage{algcompatible}
\usepackage{bm}
\usepackage{algpseudocode}


\begin{document}
\let\WriteBookmarks\relax
\def\floatpagepagefraction{1}
\def\textpagefraction{.001}

\shorttitle{ViT$^{3}$Flow for Postoperative Radiograph Synthesis in Scoliosis}
\shortauthors{Anonymous Author(s)}

\title[mode=title]{ViT$^{3}$Flow: A Test-Time Training Transformer MeanFlow for Postoperative Radiograph Synthesis in Scoliosis}

\author{
Rui Tang$^{a,*}$,
Sicheng Yang$^{c,*}$,
Moxin Zhao$^{a,*}$,
Hongqiu Wang$^{c}$,
Guankun Wang$^{e}$,
Lei Zhu$^{d}$,
Hongliang Ren$^{e}$,
Menglin Cong$^{f,\dagger}$,
and Nan Meng$^{a,b,\dagger}$
}

\affiliation{
$^{a}$ Department of Orthopaedics and Traumatology,
The University of Hong Kong,
Pokfulam, Hong Kong SAR, China.
}

\affiliation{
$^{b}$ Department of Orthopaedics and Traumatology,
The University of Hong Kong-Shenzhen Hospital,
Shenzhen 518053, China.
}

\affiliation{
$^{c}$ ROAS Thrust, Systems Hub,
The Hong Kong University of Science and Technology (Guangzhou),
Guangzhou, China.
}

\affiliation{
$^{d}$ Department of Systems Hub,
The Hong Kong University of Science and Technology (Guangzhou),
Guangzhou, China, and
The Hong Kong University of Science and Technology,
Hong Kong SAR, China.
}

\affiliation{
$^{e}$ Department of Electronic Engineering,
The Chinese University of Hong Kong,
Hong Kong SAR, China.
}

\affiliation{
$^{f}$ Department of Orthopaedic Surgery,
Qilu Hospital of Shandong University,
Jinan, Shandong 250012, China.
}

\noindent $^{*}$Rui Tang, Sicheng Yang, and Moxin Zhao contributed equally to this work.

\noindent $^{\dagger}$Corresponding authors: Menglin Cong
(\texttt{congmenglin@sdu.edu.cn}) and Nan Meng
(\texttt{nanmeng@hku.hk}).

\begin{abstract}
Predicting postoperative spinal morphology from preoperative radiographs could provide valuable support for scoliosis surgical planning, but remains challenging because surgical correction induces large spatial changes while anatomical structures must be faithfully retained. We formulate this problem as postoperative scoliosis radiograph synthesis and construct ScoliSurg, the first paired dataset for this task, comprising 632 preoperative--postoperative whole-spine radiograph pairs with structured morphology information. We further propose ViT$^{3}$Flow, a single-NFE conditional MeanFlow framework for efficient postoperative radiograph synthesis. ViT$^{3}$Flow models surgical correction as finite-interval generative transport and replaces conventional self-attention with test-time-training token mixers that perform sample-specific inner adaptation to the anatomy and deformity pattern of each case. In addition, a Spinal Morphology Extraction Agent extracts distributions of dominant-curve region and direction from the preoperative radiograph. These distributions guide Diagnosis-Routed Interval Cross-Attention (DRICA), which performs interval-dependent vertical, horizontal, joint, and global retrieval from a separate preoperative token stream. This design enables the evolving postoperative representation to incorporate spatially corresponding anatomical evidence throughout the transport process. Extensive experiments on ScoliSurg demonstrate that ViT$^{3}$Flow achieves the best performance among the compared methods in perceptual image quality, anatomical fidelity, and clinically relevant geometric accuracy, while requiring only a single network evaluation. These results highlight the potential of ViT$^{3}$Flow for efficient and anatomically faithful postoperative radiograph synthesis in scoliosis surgical planning.
\end{abstract}

\begin{highlights}
\item We formulate postoperative scoliosis radiograph synthesis as patient-conditioned finite-interval generative transport and propose ViT$^{3}$Flow, a single-NFE MeanFlow framework that integrates test-time-training-based token adaptation with diagnosis-routed retrieval of preoperative anatomy.

\item We develop the first paired dataset for postoperative scoliosis radiograph synthesis, comprising 632 preoperative--postoperative whole-spine radiograph pairs.

\item ViT$^{3}$Flow introduces a case-adaptive transport architecture that couples test-time-training token dynamics with interval-conditioned anatomical retrieval. The proposed Spinal Morphology Extraction Agent extracts distributions of dominant-curve region and direction from the preoperative radiograph, enabling Diagnosis-Routed Interval Cross-Attention (DRICA) to perform anisotropic and interval-dependent retrieval of spatially corresponding preoperative evidence throughout MeanFlow transport.

\item Extensive experiments on the proposed dataset demonstrate that ViT$^{3}$Flow consistently outperforms representative baselines in perceptual quality, anatomical fidelity, and clinically relevant geometric accuracy, highlighting its potential to support scoliosis surgical planning.
\end{highlights}


\maketitle

\input{sections/1_introduction}
\input{sections/2_related_work}
\input{sections/3_method}

\input{sections/4_experiments}
\input{sections/5_discussion}
\input{sections/6_conclusion}

\bibliographystyle{cas-model2-names}
\bibliography{references}

\end{document}

%% file: sections/1_introduction.tex
\section{Introduction}
\label{sec:introduction}

Accurate anticipation of postoperative spinal alignment is an important component of individualized scoliosis surgical planning. Preoperative whole-spine radiographs, together with Cobb angle and global-alignment measurements, provide the principal basis for characterizing deformity and defining correction objectives \cite{langensiepen2013cobb}. However, the postoperative configuration is determined by coordinated correction across multiple vertebral levels rather than by an isolated change in a single angular measurement. Its spatial pattern further depends on the preoperative curve configuration, deformity severity, and anatomy \cite{sigurdarson2025adultdeformity}. Reliable prediction of the complete postoperative spinal morphology therefore remains an unresolved requirement for surgical planning.

Current computational approaches do not directly address this requirement. Learning-based studies have primarily supported scoliosis surgical planning through preoperative deformity characterization \cite{hornung2022spinecare}, disease-progression assessment \cite{kadoury2017morphology}, and complication-risk estimation \cite{scheer2017adultdeformity}. A smaller body of work predicts postoperative angular or alignment parameters \cite{schonnagel2024lumbarfusion}. These methods provide valuable information for patient assessment, risk stratification, and quantitative outcome estimation, but describe the expected surgical result using only a limited set of numerical variables. Even when several geometric parameters are considered jointly, they cannot fully represent the spatial distribution of correction across vertebral levels or the complete postoperative appearance of the anatomy.

Deep learning-based medical image synthesis offers a potential approach for predicting postoperative radiographic appearance. Existing synthesis studies in spinal imaging have primarily focused on transformations across imaging views or modalities, including coronal-to-sagittal radiograph synthesis \cite{bassani2025sagittal}, MRI-to-CT translation \cite{graf2023mri2ct}, lumbar MRI-to-CT generation \cite{roberts2023lumbarmri2ct}, and ultrasound-to-radiograph synthesis \cite{zhou2023uxdiff}. These tasks generally preserve the underlying anatomical configuration, with the source and target images differing mainly in imaging perspective, contrast, or modality. In contrast, preoperative-to-postoperative synthesis requires the generation of a surgically altered spinal configuration while retaining patient-specific anatomical characteristics. As illustrated in Fig.~\ref{fig:intro_comparison}(a), direct image-to-image translation enables efficient conditional synthesis \cite{isola2018imagetoimagetranslationconditionaladversarial}, and registration-guided methods improve spatial correspondence \cite{kong2021reggan}; however, a direct source-to-target mapping may be insufficient to represent heterogeneous and patient-dependent structural correction. As shown in Fig.~\ref{fig:intro_comparison}(b), diffusion models provide greater generative flexibility through progressive denoising \cite{ho2020denoisingdiffusionprobabilisticmodels}, whereas flow matching formulates generation as continuous probability transport \cite{lipman2023flowmatchinggenerativemodeling}. These approaches commonly rely on iterative sampling with multiple network evaluations, increasing latency and computational cost in clinical planning workflows. This motivates a low-NFE formulation capable of modeling large corrections while preserving anatomical correspondence.

MeanFlow provides an efficient formulation for this task by learning an interval-averaged velocity rather than only an instantaneous velocity \cite{geng2025meanflowsonestepgenerative}, allowing a single network evaluation to represent a substantial portion of the generative trajectory. However, MeanFlow primarily reduces temporal discretization and does not by itself determine how conditions should be represented or incorporated into the transport field. The predicted interval velocity is still parameterized by a generative backbone, and a backbone with token-mixing parameters shared across all samples applies the same learned token-processing function to scoliosis cases with markedly different vertebral morphology, dominant-curve region, curve direction, and deformity pattern. Moreover, treating the preoperative radiograph as a global condition does not distinguish regions requiring substantial correction from structures whose anatomical identity should be retained. Because each finite-interval update covers a large portion of the trajectory, inadequate sample-specific adaptation or imprecise source conditioning may propagate structural errors over the entire interval rather than allowing them to be progressively corrected through multiple local evaluations. Accordingly, the effectiveness of finite-interval transport depends not only on the transport formulation itself, but also on the model's ability to accommodate inter-patient variation and maintain anatomically grounded conditioning.

To address these challenges, we construct \textbf{ScoliSurg}, to our knowledge the first paired dataset for postoperative scoliosis radiograph synthesis, comprising 632 matched preoperative and postoperative radiograph pairs. ScoliSurg enables the direct learning and evaluation of postoperative morphology. As illustrated in Fig.~\ref{fig:intro_comparison}(c), we further propose \textbf{ViT$^{3}$Flow}, a single-NFE generative framework that models postoperative synthesis as patient-conditioned finite-interval transport. ViT$^{3}$Flow parameterizes the interval-averaged velocity using a DiT-style backbone \cite{peebles2023scalablediffusionmodelstransformers}, in which visual test-time-training token mixers \cite{han2026vit3unlockingtesttimetraining} perform sample-specific inner updates to adapt the token-mixing computation to the current case and generative state. To introduce structured preoperative conditioning, the proposed \textbf{Spinal Morphology Extraction Agent (SEMA)} estimates probability distributions over the dominant-curve region and image-coordinate direction, together with their joint configuration, from each preoperative radiograph. These morphology conditions guide \textbf{Diagnosis Routed Interval Cross Attention (DRICA)} in performing anisotropic and interval-dependent retrieval from a separate preoperative token stream, enabling the evolving postoperative representation to incorporate spatially organized anatomical evidence during finite-interval transport. The main contributions of this work are summarized as follows:

\begin{figure}[pos=t]
\centering
\includegraphics[width=\columnwidth]{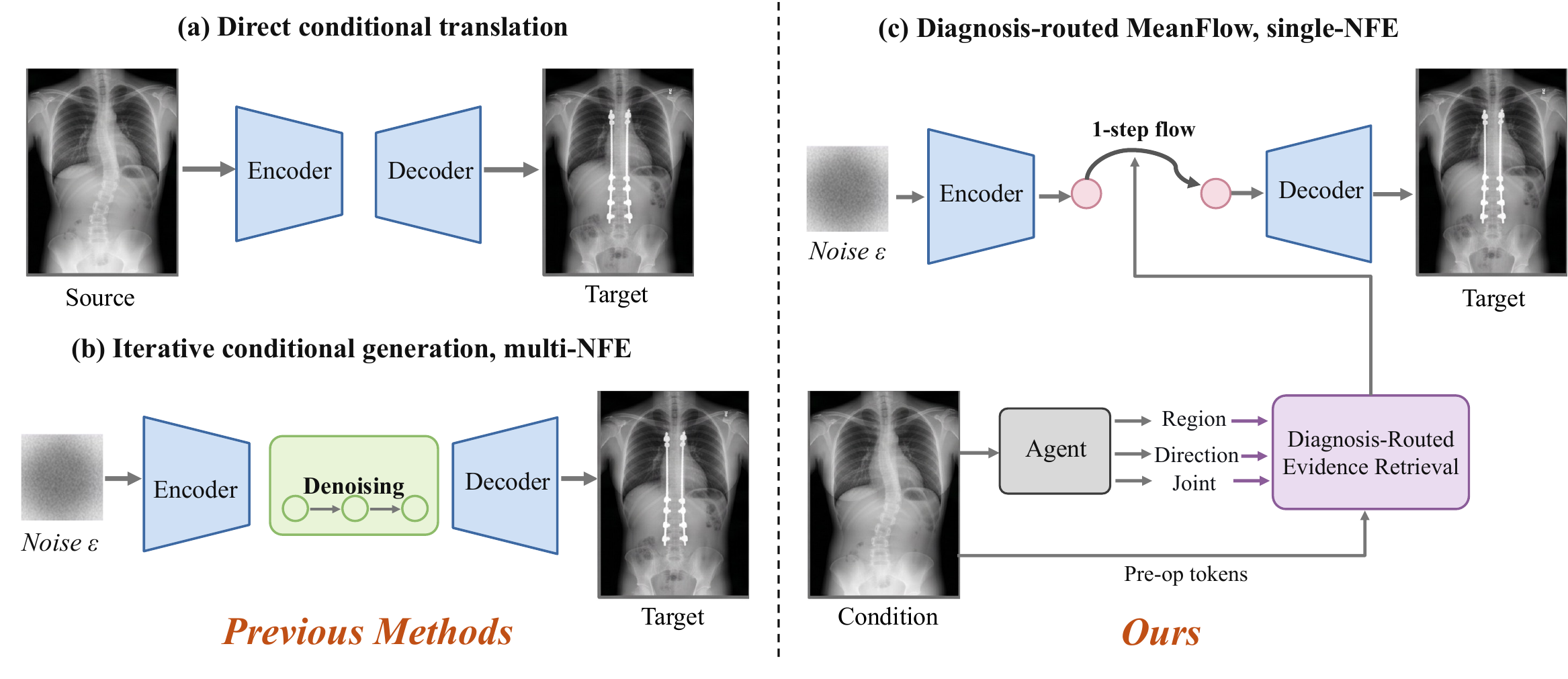}
\vspace{-2mm}
\caption{Comparisons of generative paradigms for preoperative-to-postoperative scoliosis radiograph synthesis. Direct conditional translation (a) predicts the postoperative radiograph through a single feed-forward mapping, while iterative conditional generation (b) progressively transforms noise into the target image using multiple network function evaluations. Our ViT$^{3}$Flow (c) performs patient-conditioned finite-interval generation with a single network evaluation, where the Spinal Morphology Extraction Agent provides dominant-curve region and direction distributions to guide Diagnosis-Routed Interval Cross-Attention in retrieving spatially organized evidence from the preoperative radiograph.
}
\vspace{-3mm}
\label{fig:intro_comparison}
\end{figure}

\begin{itemize}
\item We formulate postoperative scoliosis radiograph synthesis as patient-conditioned finite-interval generative transport and propose \textbf{ViT$^{3}$Flow}, a single-NFE MeanFlow framework that integrates sample-specific token-mixing adaptation with diagnosis-routed retrieval of preoperative anatomy.

\item We construct ScoliSurg, to our knowledge the first paired dataset for postoperative scoliosis radiograph synthesis, comprising 632 matched preoperative and postoperative whole-spine radiograph pairs with structured morphology information for model development and evaluation.

\item We introduce the Spinal Morphology Extraction Agent to estimate probability distributions over the dominant-curve region and image-coordinate direction from preoperative radiographs. These distributions and their joint configuration enable Diagnosis Routed Interval Cross Attention (DRICA) to perform anisotropic and interval-dependent retrieval of spatially organized preoperative evidence.

\item Extensive experiments on ScoliSurg demonstrate that ViT$^{3}$Flow outperforms representative baselines in perceptual image quality, anatomical fidelity, and clinically relevant geometric accuracy while requiring only a single network evaluation.
\end{itemize}

%% file: sections/2_related_work.tex
\section{Related Work}
\label{sec:related}

\subsection{Scoliosis Surgical Planning and Postoperative Prediction}

Artificial intelligence (AI) has been increasingly applied to scoliosis assessment, with most studies focusing on the deformity visible in the current radiograph, including curve characterization, severity assessment, and morphological classification \cite{hornung2022spinecare,kadoury2017morphology}. Subsequent work has extended these applications to disease-progression assessment, complication-risk estimation \cite{scheer2017adultdeformity}, and prediction of selected postoperative angular or alignment measurements \cite{schonnagel2024lumbarfusion,sigurdarson2025adultdeformity}. These methods provide valuable information about the current deformity and possible clinical outcomes, but their predictions are generally expressed as classifications, probabilities, or a limited set of geometric measurements. Such outputs cannot directly represent the spatial distribution of surgical correction across vertebral levels or the complete postoperative spinal configuration.

To provide a more spatially resolved representation, generative modeling has been explored in spinal imaging. Existing studies, however, have mainly addressed transformations across imaging views or modalities, including coronal-to-sagittal radiograph synthesis \cite{bassani2025sagittal}, MRI-to-CT translation for spinal analysis \cite{graf2023mri2ct}, lumbar MRI-to-CT generation \cite{roberts2023lumbarmri2ct}, and ultrasound-to-radiograph synthesis \cite{zhou2023uxdiff}. These tasks primarily alter imaging perspective, modality, or tissue contrast while generally preserving the underlying anatomical configuration. Preoperative-to-postoperative synthesis differs because it must model a surgically induced change in spinal alignment while retaining correspondence with the anatomy visible in the preoperative radiograph. Paired preoperative--postoperative data and generative studies specifically developed for postoperative scoliosis radiograph synthesis remain limited.

\subsection{Generative Medical Image Translation}

Medical image synthesis is commonly formulated as conditional image-to-image translation \cite{isola2018imagetoimagetranslationconditionaladversarial,wang2024non,yu2024rethinking}. Pix2Pix learns paired source-to-target mappings through adversarial and reconstruction objectives \cite{isola2018imagetoimagetranslationconditionaladversarial}, whereas CycleGAN enables unpaired translation through cycle-consistency constraints \cite{zhu2020unpairedimagetoimagetranslationusing}. Registration-guided and transformer-based approaches further improve spatial correspondence and long-range feature interaction in multimodal medical translation \cite{kong2021reggan,dalmaz2022resvit}. These methods provide efficient direct inference, but a single feed-forward mapping may be restrictive when the target involves large and spatially heterogeneous anatomical changes. Diffusion models offer a probabilistic alternative by progressively reversing a predefined noising process \cite{ho2020denoisingdiffusionprobabilisticmodels}, while image-to-image Schr\"odinger bridges learn stochastic transitions between source and target distributions \cite{liu2023i2sb}. Although these formulations provide greater flexibility for conditional synthesis, their iterative sampling generally requires repeated network evaluations.

Flow matching formulates generation as continuous probability transport by learning an instantaneous velocity field along a prescribed probability path \cite{lipman2023flowmatchinggenerativemodeling}. Samples are generated by numerically integrating this field from the initial distribution toward the target distribution. Optimal-transport flow matching shortens the transport path and has been explored for medical image synthesis across different modalities, spatial dimensions, and conditioning settings \cite{yazdani2025flow}; however, accurate generation generally remains dependent on multiple evaluations of the instantaneous velocity field. MeanFlow instead learns an average velocity over a finite interval, allowing a nonzero portion of the generative trajectory to be represented within a single network evaluation and enabling single-NFE generation without distillation or a pretrained teacher \cite{geng2025meanflowsonestepgenerative}. This formulation provides a promising basis for improving the efficiency of medical image synthesis, but its application to conditional medical image translation remains insufficiently explored.

\subsection{Test-Time Training in Visual Modeling}

Transformer-based backbones have become central to visual generation because self-attention enables global interactions among spatial tokens \cite{dosovitskiy2021vit}. Diffusion Transformers (DiTs) further demonstrate that transformer blocks can effectively parameterize the denoising or velocity fields of generative models \cite{peebles2023scalablediffusionmodelstransformers}. In conventional DiT backbones, attention weights vary with the input, but the projection and token-mixing parameters remain fixed after training and are shared across all samples at inference time. The resulting computation is therefore input dependent at the feature level but not explicitly adapted at the parameter level to the current case. This distinction is important for postoperative synthesis, where spinal morphology, deformity configuration, and the required correction pattern can vary substantially across individuals. Under single-NFE generation, such variation must be accommodated within one evaluation of the generative backbone rather than through repeated local refinement.

Test-Time Training (TTT) was originally introduced as a self-supervised adaptation strategy that updates a model on each unlabeled test sample before prediction \cite{sun2020testtime}. Subsequent work extended this idea in visual recognition by using masked autoencoding as the one-sample test-time objective, improving robustness under distribution shifts \cite{gandelsman2022testtime}. Building on this adaptation perspective, TTT reformulates token mixing as a sample-dependent online learning process. The key--value tokens of the current input are treated as an inner training set and used to update the parameters of a compact inner model, after which the adapted model is applied to the corresponding queries. ViT$^{3}$ systematically develops this formulation for visual modeling through dedicated inner-training objectives and inner-model architectures, and demonstrates its effectiveness across image classification, generation, object detection, and semantic segmentation \cite{han2026vit3unlockingtesttimetraining}. Because the inner parameters are updated separately for each visual sequence, ViT$^{3}$ allows the token-processing function itself to specialize to the spatial structure of the current sample while retaining a shared outer model. This sample-specific adaptation provides a suitable mechanism for modeling inter-patient variation within a generative backbone.

%% file: sections/3_method.tex
\section{Method}
\label{sec:method}

\subsection{Overview}
\label{subsec:overview}

Postoperative scoliosis radiograph synthesis involves large and spatially heterogeneous corrective changes, making it challenging to model postoperative morphology while maintaining correspondence with patient-specific preoperative anatomy. To enable efficient generation without relying on iterative refinement, ViT$^{3}$Flow formulates the transformation as a finite-interval MeanFlow process, allowing the postoperative state to be generated with a single network function evaluation (NFE).

\begin{figure*}[pos=t]
\centering
\includegraphics[width=\textwidth]{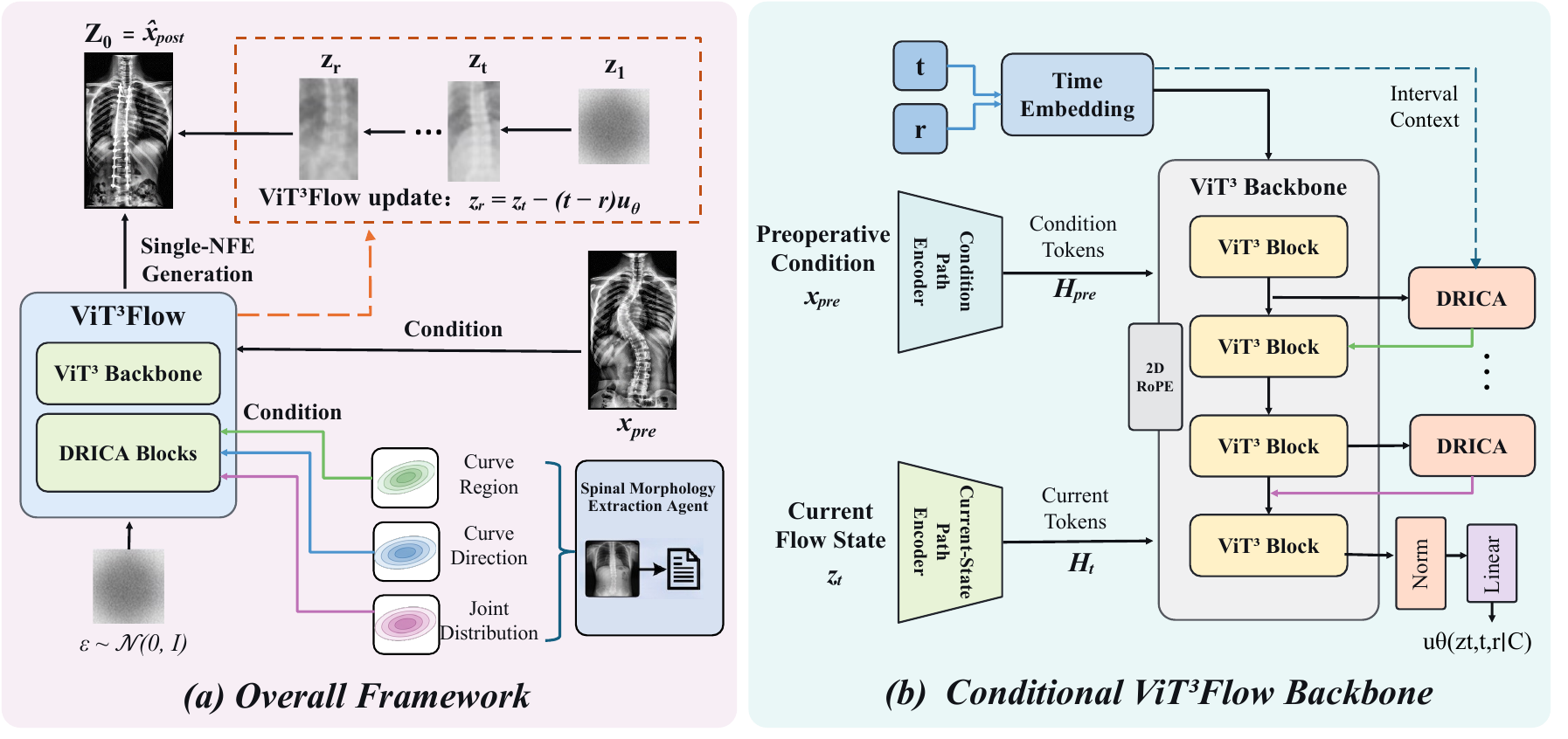}
\vspace{-2mm}
\caption{Overview of the proposed ViT$^{3}$Flow framework. (a) The preoperative radiograph $x_{\mathrm{pre}}$ provides the image condition and is analyzed by SMEA to obtain the dominant-curve region probability vector $\boldsymbol{q}_{\mathrm{reg}}$, image-coordinate curve-direction probability vector $\boldsymbol{q}_{\mathrm{dir}}$, and joint region--direction distribution $\boldsymbol{q}_{\mathrm{joint}}$. Starting from $z_1\sim\mathcal{N}(0,I)$, ViT$^{3}$Flow predicts an interval-averaged velocity and synthesizes the postoperative radiograph under the principal single-NFE setting. The dashed box illustrates the general finite-interval update $z_r=z_t-(t-r)u_{\theta}$. (b) The preoperative radiograph and current flow state are independently encoded into preoperative tokens $H_{\mathrm{pre}}$ and current tokens $H_t$. Both streams are supplied to the unified ViT$^{3}$Flow backbone: $H_t$ is sequentially processed by the ViT$^{3}$ blocks, whereas the fixed $H_{\mathrm{pre}}$ is accessed by DRICA at selected depths. The interval embeddings condition the ViT$^{3}$ blocks and provide interval context for DRICA, whose outputs update the evolving current representation before the conditional interval-averaged velocity $u_{\theta}(z_t,t,r\mid\mathcal{C})$ is predicted.}
\vspace{-3mm}
\label{fig:framework}
\end{figure*}

As illustrated in Fig.~\ref{fig:framework}, given a preoperative radiograph $x_{\mathrm{pre}}\in[0,1]^{1\times H\times W}$, ViT$^{3}$Flow generates the corresponding postoperative radiograph $\hat{x}_{\mathrm{post}}$ from Gaussian noise through a patient-conditioned finite-interval process. The Spinal Morphology Extraction Agent (SMEA) first analyzes $x_{\mathrm{pre}}$ to obtain a dominant-curve region probability vector $\boldsymbol{q}_{\mathrm{reg}}$, an image-coordinate curve-direction probability vector $\boldsymbol{q}_{\mathrm{dir}}$, and their joint distribution $\boldsymbol{q}_{\mathrm{joint}}$. During generation, the ViT$^{3}$ backbone adapts token computation to the current case, while Diagnosis Routed Interval Cross Attention (DRICA) uses the morphology information together with the current interval context to retrieve spatial evidence from the preoperative radiograph and inject it into the evolving postoperative representation. The complete condition is defined as
\begin{equation}
\mathcal{C}
=
\left(
x_{\mathrm{pre}},
\boldsymbol{q}_{\mathrm{reg}},
\boldsymbol{q}_{\mathrm{dir}},
\boldsymbol{q}_{\mathrm{joint}}
\right),
\label{eq:complete_condition}
\end{equation}
under which the model learns the conditional postoperative distribution $p_{\theta}(x_{\mathrm{post}}\mid\mathcal{C})$.

Generation starts from Gaussian noise $z_1=\epsilon$, where $\epsilon\sim\mathcal{N}(0,I)$. Given the current flow state $z_t$, interval endpoints $(t,r)$, and condition $\mathcal{C}$, ViT$^{3}$Flow predicts the conditional interval-averaged velocity $u_{\theta}(z_t,t,r\mid\mathcal{C})$ and updates the state as
\begin{equation}
z_r
=
z_t
-
(t-r)
u_{\theta}
\left(
z_t,t,r\mid\mathcal{C}
\right),
\qquad
\label{eq:overview_update}
\end{equation}
Here $0\leq r\leq t\leq1.$.The case $r=t$ corresponds to the instantaneous flow-matching boundary, whereas $r<t$ corresponds to finite-interval generation. In the principal single-NFE setting, the complete transition is evaluated over $(t,r)=(1,0)$, producing the terminal state $z_0$ with one network evaluation. The final postoperative prediction $\hat{x}_{\mathrm{post}}$ is obtained by denormalizing and clipping $z_0$ to the valid image range. The same formulation also supports optional few-step inference by partitioning the complete trajectory into multiple finite intervals.

Figure~\ref{fig:framework}(b) illustrates the conditional ViT$^{3}$Flow backbone. The preoperative radiograph $x_{\mathrm{pre}}$ and the current flow state $z_t$ are independently encoded by the condition path and current-state path to form the preoperative tokens $H_{\mathrm{pre}}$ and current tokens $H_t$, respectively. Their complete definitions, including the corresponding positional embeddings, are provided in Sec.~\ref{subsec:vit3flow}. The two token streams enter the unified backbone but serve different roles. The current tokens are sequentially processed by the ViT$^{3}$ blocks to represent the evolving postoperative state, whereas the preoperative tokens retain the spatial organization of the source anatomy and remain fixed throughout the forward pass. Within each ViT$^{3}$ block, test-time training performs a sample-specific inner update on the current token sequence, allowing token computation to adapt to the current case. Two-dimensional rotary positional encoding further preserves the vertical and horizontal organization of the whole-spine token grid.

The interval endpoints $t$ and $r$ are independently embedded, and the resulting interval representations condition the adaptive normalization layers of the ViT$^{3}$ blocks and provide the interval context used by DRICA. Their formal definition is provided in Sec.~\ref{subsec:vit3flow}.

At selected backbone depths, the intermediate current tokens are passed to DRICA, where they act as queries and the fixed preoperative tokens $H_{\mathrm{pre}}$ provide keys and values. The region probabilities $\boldsymbol{q}_{\mathrm{reg}}$ guide retrieval along the vertical image axis, the direction probabilities $\boldsymbol{q}_{\mathrm{dir}}$ guide retrieval along the horizontal axis, and the joint distribution $\boldsymbol{q}_{\mathrm{joint}}$ regulates cross-axis interaction. Global Context Retrieval further provides access to information across the complete preoperative token sequence without applying the morphology-derived spatial biases. The retrieved evidence is fused and written back to the current stream before processing continues through the subsequent ViT$^{3}$ blocks. After the final block, an interval-conditioned prediction head projects the current tokens into the conditional interval-averaged velocity $u_{\theta}(z_t,t,r\mid\mathcal{C})$, which is used in the finite-interval update to obtain the state at the target endpoint.

\subsection{Spinal Morphology Extraction Agent}
\label{subsec:smea}

Although the preoperative radiograph contains rich patient-specific anatomical information, the image condition alone does not explicitly identify the location and orientation of the dominant spinal curve, making morphology-aware retrieval less direct during generation. To provide structured patient-specific cues, the Spinal Morphology Extraction Agent (SMEA) analyzes only the preoperative frontal radiograph and predicts the dominant-curve region and its direction in image coordinates. The region is categorized as thoracic, thoracolumbar, or lumbar, whereas the direction is categorized as image-left or image-right according to the convexity of the dominant curve. The fixed prompt restricts the analysis to these visible preoperative attributes, requests a structured JSON response, and prohibits inference of surgical procedures or postoperative outcomes.

For the full dataset, each radiograph was initially queried $K=7$ times using GPT-5.5. Each query was performed as a single-turn multimodal interaction with a temperature of $0.5$ and a maximum output length of 120 tokens. The valid responses were aggregated into a dominant-curve region probability vector $\boldsymbol{q}_{i}^{\mathrm{reg}}$, an image-coordinate curve-direction probability vector $\boldsymbol{q}_{i}^{\mathrm{dir}}$, and their joint region-direction distribution $\boldsymbol{q}_{i}^{\mathrm{joint}}$ for case $i$. The querying, parsing, and aggregation procedure is summarized in Algorithm~\ref{alg:smea}.

For the $k$-th valid response, let $a_i^{(k)}=(\ell_i^{(k)},d_i^{(k)})$, where $\ell_i^{(k)}$ and $d_i^{(k)}$ denote the predicted region and direction, respectively. Let $\mathcal{V}_i$ denote the set of valid response indices obtained from the initial queries and, when necessary, at most $R$ additional retries. When $\mathcal{V}_i$ is nonempty, the joint region--direction distribution is computed as
\begin{equation}
q_{i,c,m}^{\mathrm{joint}}
=
\frac{1}{|\mathcal{V}_i|}
\sum_{k\in\mathcal{V}_i}
\mathbb{I}
\left[
\ell_i^{(k)}=c,\,
d_i^{(k)}=m
\right],
\label{eq:joint_morphology_distribution}
\end{equation}
where $c\in\{1,2,3\}$ indexes the thoracic, thoracolumbar, and lumbar regions, and $m\in\{1,2\}$ indexes the image-left and image-right directions. If no valid response is obtained after the finite retry budget, $\boldsymbol{q}_{i}^{\mathrm{joint}}$ is set to the uniform distribution over the six region--direction combinations. The corresponding marginal probabilities are obtained as
\begin{equation}
q_{i,c}^{\mathrm{reg}}
=
\sum_{m=1}^{2}
q_{i,c,m}^{\mathrm{joint}},
\qquad
q_{i,m}^{\mathrm{dir}}
=
\sum_{c=1}^{3}
q_{i,c,m}^{\mathrm{joint}}.
\label{eq:marginal_morphology_probabilities}
\end{equation}
We denote the resulting morphology conditions by $\boldsymbol{q}_i=(\boldsymbol{q}_{i}^{\mathrm{reg}},\boldsymbol{q}_{i}^{\mathrm{dir}},\boldsymbol{q}_{i}^{\mathrm{joint}})$. These conditions are computed once for each preoperative radiograph and subsequently used to route source-evidence retrieval in DRICA.

\subsection{ViT$^{3}$Flow for Finite-Interval Generation}
\label{subsec:vit3flow}

To address the reliance of instantaneous-velocity models on repeated network evaluations, ViT$^{3}$Flow adopts a finite-interval MeanFlow formulation. Let $y=\operatorname{norm}(x_{\mathrm{post}})$ denote the normalized postoperative target and let $\epsilon\sim\mathcal{N}(0,I)$. We construct a linear probability path between the postoperative target and Gaussian noise as
\begin{equation}
z_t=(1-t)y+t\epsilon,
\qquad
v_t=\frac{\mathrm{d}z_t}{\mathrm{d}t}
=\epsilon-y.
\label{eq:probability_path}
\end{equation}
Here, $t\in[0,1]$. Accordingly, $z_0=y$ corresponds to the postoperative target, $z_1=\epsilon$ corresponds to Gaussian noise, and $v_t$ denotes the instantaneous velocity along the linear path.

ViT$^{3}$Flow predicts the conditional interval-averaged velocity $u_{\theta}(z_t,t,r\mid\mathcal{C})$ over $[r,t]$, where $0\leq r\leq t\leq1$. The state at the second endpoint is obtained through
\begin{equation}
z_r
=
z_t
-
(t-r)
u_{\theta}
\left(
z_t,t,r\mid\mathcal{C}
\right).
\label{eq:interval_update}
\end{equation}
For $r<t$, each prediction represents the evolution across a nonzero interval rather than only the local change at a single time point, whereas $r=t$ corresponds to the instantaneous flow-matching boundary. Setting $(t,r)=(1,0)$ directly maps the initial noise state to the postoperative state with a single network function evaluation (NFE), whereas recursively applying Eq.~\eqref{eq:interval_update} over a partitioned schedule supports optional few-step inference.

We parameterize the interval-averaged velocity using separate current-state and preoperative token streams:
\begin{equation}
H_t^{0}
=
E_{\mathrm{cur}}(z_t)+P_{\mathrm{cur}},
\qquad
H_{\mathrm{pre}}
=
E_{\mathrm{pre}}(x_{\mathrm{pre}})+P_{\mathrm{pre}},
\label{eq:dual_stream_encoding}
\end{equation}
where $E_{\mathrm{cur}}$ and $E_{\mathrm{pre}}$ are independent patch encoders, and $P_{\mathrm{cur}}$ and $P_{\mathrm{pre}}$ are fixed two-dimensional positional embeddings. The current tokens $H_t^{0}$ are progressively transformed to represent the evolving postoperative state, whereas the preoperative tokens $H_{\mathrm{pre}}$ remain fixed and retain the spatial organization of the source anatomy.

The two interval endpoints are independently embedded as
\begin{equation}
e_t=E_t(t),
\qquad
e_r=E_r(r).
\label{eq:endpoint_embeddings}
\end{equation}
Their sum and difference are then used to represent the current finite interval:
\begin{equation}
p_{r,t}=e_t+e_r,
\qquad
d_{r,t}=e_t-e_r.
\label{eq:interval_embeddings}
\end{equation}
The summed representation $p_{r,t}$ conditions the adaptive normalization layers of the ViT$^{3}$ backbone, while the pair $(p_{r,t},d_{r,t})$ is further encoded as the interval context supplied to DRICA.

Finite-interval generation assigns a larger portion of the noise-to-image transformation to each network evaluation, increasing the need for the backbone to adapt its computation to the morphology and current generative state of an individual case. However, conventional self-attention applies the same learned token-mixing parameters to all samples. To introduce sample-specific computation within each evaluation, the current token stream is processed by interval-conditioned ViT$^{3}$ blocks in which self-attention is replaced by a visual test-time-training (TTT) mixer.

For input sample $i$, the key and value representations of the current token sequence form an inner training set. A compact inner model is updated using these representations and then applied to the corresponding queries:
\begin{equation}
\phi_i'
=
\phi
-
\eta
\nabla_{\phi}
\mathcal{L}_{\mathrm{TTT}}
\left(
K_i,V_i;\phi
\right),
\qquad
\widetilde{H}_i
=
f_{\phi_i'}(Q_i),
\label{eq:ttt_update}
\end{equation}
where $\phi$ denotes the shared initialization of the inner model and $\eta$ is the inner-update rate. The adapted parameters $\phi_i'$ are used only for the current input and are reset to $\phi$ before processing the next sample. Consequently, the token-processing function is specialized to the current case without permanently modifying the shared backbone parameters.

The TTT mixer contains a simplified SwiGLU branch for adaptive channel-wise transformation and a $3\times3$ depth-wise convolution branch for local spatial modeling. Both branches are updated using the current key--value sequence and subsequently evaluated on the queries. Their outputs are concatenated and projected to the original token dimension. Two-dimensional rotary positional encoding is applied to the query and key representations of the SwiGLU branch to preserve the vertical and horizontal organization of the whole-spine token grid.

At the $\ell$-th backbone layer, the current representation is updated as
\begin{equation}
\bar{H}_t^{\ell}
=
\mathcal{B}_{\ell}
\left(
H_t^{\ell-1};
p_{r,t}
\right),
\label{eq:vit3_backbone_update}
\end{equation}
where $\mathcal{B}_{\ell}$ denotes an interval-conditioned ViT$^{3}$ block. At selected depths, the intermediate representation $\bar{H}_t^{\ell}$ is further updated by DRICA using the fixed preoperative tokens $H_{\mathrm{pre}}$, the interval context, and the SMEA-derived morphology conditions. The detailed source-retrieval mechanism is presented in Sec.~\ref{subsec:drica}.

After the final backbone layer, an interval-conditioned prediction head maps the current tokens to the interval-averaged velocity:
\begin{equation}
u_{\theta}
\left(
z_t,t,r\mid\mathcal{C}
\right)
=
\operatorname{Unpatchify}
\left[
\operatorname{Head}
\left(
H_t^{L};
p_{r,t}
\right)
\right].
\label{eq:velocity_output}
\end{equation}
The prediction head applies interval-conditioned normalization followed by linear projection, and $\operatorname{Unpatchify}(\cdot)$ restores the token sequence to the spatial image layout. The resulting velocity has the same dimensionality as $z_t$ and is substituted into Eq.~\eqref{eq:interval_update} to obtain the state at endpoint $r$. Under single-NFE inference, the model evaluates $u_{\theta}(z_1,1,0\mid\mathcal{C})$ once and obtains the postoperative state as $z_0=z_1-u_{\theta}(z_1,1,0\mid\mathcal{C})$.

\begin{algorithm}[H]
\caption{SMEA extraction of morphology-based routing information}
\label{alg:smea}
\begin{algorithmic}[1]
\Require Preoperative radiograph $x_{\mathrm{pre}}^{(i)}$, multimodal model
$\mathcal{M}$, fixed prompt $\mathcal{P}$, initial query number $K$, retry limit $R$
\Ensure Morphological distributions
$\boldsymbol{q}_{i}^{\mathrm{reg}}$,
$\boldsymbol{q}_{i}^{\mathrm{dir}}$, and
$\boldsymbol{q}_{i}^{\mathrm{joint}}$

\State $\mathcal{L}\gets
\{\texttt{thoracic},\texttt{thoracolumbar},\texttt{lumbar}\}$
\State $\mathcal{D}\gets
\{\texttt{image\_left},\texttt{image\_right}\}$
\State $\mathcal{A}_i\gets[\,]$

\For{$k=1,\ldots,K$}
    \State $y_i^{(k)}\gets
    \Call{Query}{\mathcal{M},\mathcal{P},x_{\mathrm{pre}}^{(i)}}$
    \State $(\ell_i^{(k)},d_i^{(k)})\gets
    \Call{ParseResponse}{y_i^{(k)}}$
    \If{$\ell_i^{(k)}\in\mathcal{L}$ \textbf{and}
        $d_i^{(k)}\in\mathcal{D}$}
        \State \Call{Append}{$\mathcal{A}_i,
        (\ell_i^{(k)},d_i^{(k)})$}
    \EndIf
\EndFor

\State $r\gets0$
\While{$\mathcal{A}_i=[\,]$ \textbf{and} $r<R$}
    \State $y\gets\Call{Query}
    {\mathcal{M},\mathcal{P},x_{\mathrm{pre}}^{(i)}}$
    \State $(\ell,d)\gets\Call{ParseResponse}{y}$
    \If{$\ell\in\mathcal{L}$ \textbf{and} $d\in\mathcal{D}$}
        \State \Call{Append}{$\mathcal{A}_i,(\ell,d)$}
    \EndIf
    \State $r\gets r+1$
\EndWhile

\If{$\mathcal{A}_i=[\,]$}
    \State $\boldsymbol{q}_{i}^{\mathrm{joint}}
    \gets\Call{UniformDistribution}{|\mathcal{L}|\times|\mathcal{D}|}$
\Else
    \State $\boldsymbol{q}_{i}^{\mathrm{joint}}
    \gets\Call{NormalizedJointHistogram}{\mathcal{A}_i}$
\EndIf

\State $\boldsymbol{q}_{i}^{\mathrm{reg}}
\gets\sum_{d\in\mathcal{D}}\boldsymbol{q}_{i}^{\mathrm{joint}}(:,d)$
\State $\boldsymbol{q}_{i}^{\mathrm{dir}}
\gets\sum_{\ell\in\mathcal{L}}\boldsymbol{q}_{i}^{\mathrm{joint}}(\ell,:)$
\State \Return
$\boldsymbol{q}_{i}^{\mathrm{reg}},
\boldsymbol{q}_{i}^{\mathrm{dir}},
\boldsymbol{q}_{i}^{\mathrm{joint}}$
\end{algorithmic}
\end{algorithm}

\subsection{Diagnosis Routed Interval Cross Attention}
\label{subsec:drica}

Although finite-interval MeanFlow enables low-NFE generation and the ViT$^{3}$ backbone adapts token computation to the current case, these mechanisms do not explicitly determine how the evolving representation should retrieve spatial evidence from the corresponding preoperative radiograph. Conventional global cross attention primarily relies on feature similarity and does not organize source retrieval according to the dominant-curve morphology or the current finite interval. To address this remaining limitation, we propose Diagnosis-Routed Interval Cross Attention (DRICA), which introduces morphology- and interval-dependent source retrieval into the ViT$^{3}$ backbone.

\begin{figure*}[pos=t]
\centering
\includegraphics[width=\textwidth]{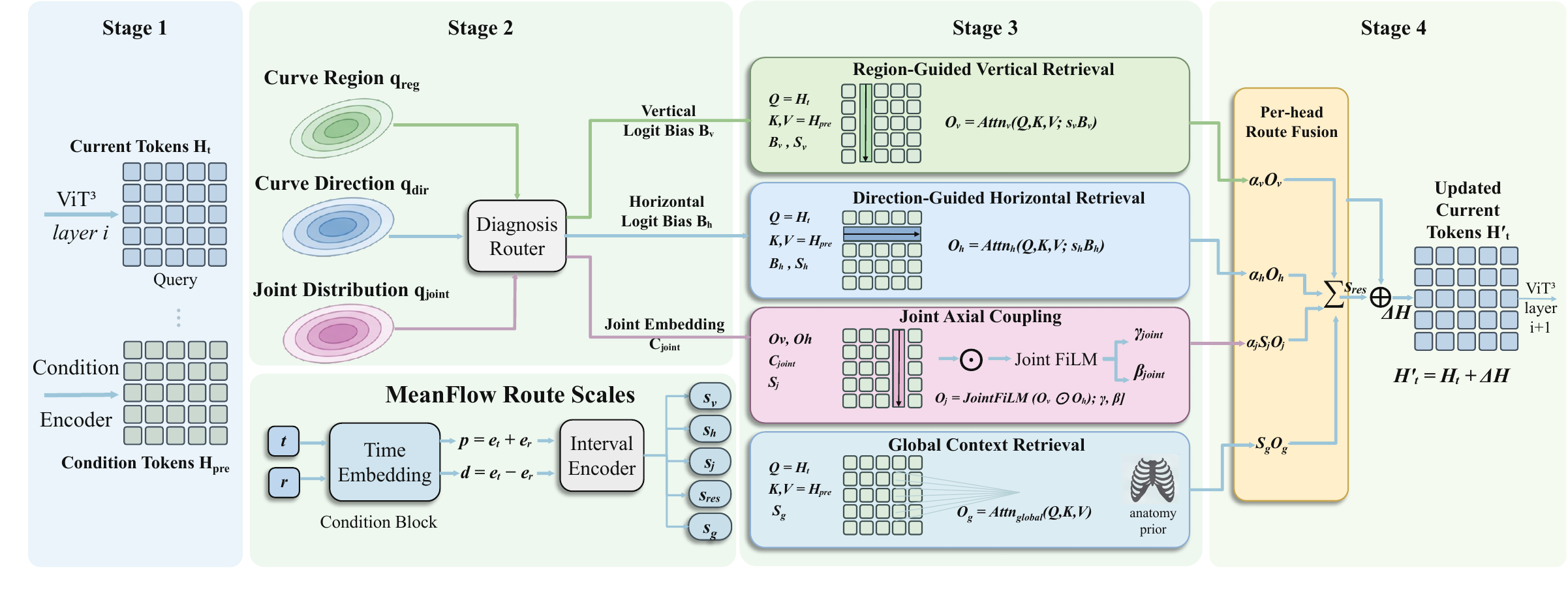}
\caption{Overview of Diagnosis Routed Interval Cross Attention (DRICA). The evolving current tokens $H_t$ retrieve evidence from the fixed preoperative tokens $H_{\mathrm{pre}}$ through region-guided vertical retrieval, direction-guided horizontal retrieval, joint cross-axis coupling, and diagnosis-independent global retrieval. The MeanFlow interval context controls the route scales and residual write-back before producing the updated current tokens $H_t'$.}
\label{fig:drica}
\end{figure*}

As illustrated in Fig.~\ref{fig:drica}, DRICA receives the evolving current tokens $H_t$ and the fixed preoperative tokens $H_{\mathrm{pre}}$. The current tokens are used to construct queries, whereas the preoperative tokens provide keys and values. DRICA then constructs morphology-based and interval dependent routing controls, applies them to four complementary source-retrieval pathways, and writes the fused evidence back to the current representation before subsequent ViT$^{3}$ processing.

DRICA first constructs two complementary sets of routing controls. The diagnosis router converts the SMEA-derived region probabilities, direction probabilities, and joint region direction distribution into spatial routing information. In parallel, the interval encoder transforms the MeanFlow endpoint representations $(p_{r,t},d_{r,t})$ into interval-dependent scales that regulate the retrieval pathways and residual write-back. Using the interval representations defined in Eq.~\eqref{eq:interval_embeddings}, the interval context is encoded as
\begin{equation}
h_{r,t}
=
E_I
\left(
[p_{r,t};d_{r,t}]
\right).
\label{eq:drica_interval_context}
\end{equation}
The context $h_{r,t}$ represents both the temporal positions of the two endpoints and their separation. For each attention head $h$, the interval encoder predicts five route scales:
\begin{equation}
\left(
s_v^h,
s_h^h,
s_j^h,
s_{\mathrm{res}}^h,
s_g^h
\right)
=
\sigma
\left(
W_I^h h_{r,t}
\right),
\label{eq:interval_route_scales}
\end{equation}
where $s_v^h$ and $s_h^h$ regulate the vertical and horizontal routing biases, $s_j^h$ controls the joint axial pathway, $s_{\mathrm{res}}^h$ controls the overall residual write-back, and $s_g^h$ regulates the Global Context Retrieval pathway. Consequently, the contribution of the fixed morphology conditions varies according to the current MeanFlow interval.

The diagnosis router converts the dominant-curve region and direction probabilities into head-specific spatial bias profiles:
\begin{equation}
B_v^h
=
\sum_{c=1}^{3}
q_{i,c}^{\mathrm{reg}}
\mathcal{B}_{v,c}^h,
\qquad
B_h^h
=
\sum_{m=1}^{2}
q_{i,m}^{\mathrm{dir}}
\mathcal{B}_{h,m}^h,
\label{eq:morphology_bias_profiles}
\end{equation}
where $\mathcal{B}_{v,c}^h$ and $\mathcal{B}_{h,m}^h$ are learnable bias profiles associated with the vertical curve regions and horizontal curve directions, respectively. Their probability-weighted combinations provide continuous morphology-dependent routing biases rather than selecting a single hard category. The joint region--direction distribution is separately encoded as
\begin{equation}
c_{\mathrm{joint}}
=
E_J
\left(
\boldsymbol{q}_{i}^{\mathrm{joint}}
\right),
\label{eq:joint_morphology_embedding}
\end{equation}
which provides the joint morphology representation used for cross-axis interaction and per-head route fusion.

Using the morphology derived biases and interval-dependent scales, DRICA performs Region-Guided Vertical Retrieval and Direction-Guided Horizontal Retrieval:
\begin{equation}
\begin{aligned}
O_v^h
&=
\operatorname{softmax}
\left[
\frac{Q_v^h(K_v^h)^{\top}}{\sqrt{d_h}}
+
s_v^h B_v^h
\right]V_v^h,
\\
O_h^h
&=
\operatorname{softmax}
\left[
\frac{Q_h^h(K_h^h)^{\top}}{\sqrt{d_h}}
+
s_h^h B_h^h
\right]V_h^h.
\end{aligned}
\label{eq:axial_retrieval}
\end{equation}
Here, the query representations are derived from $H_t$, whereas the key and value representations are derived from $H_{\mathrm{pre}}$. Vertical retrieval is performed independently within each image column, while horizontal retrieval is performed independently within each image row. Because the morphology profiles are introduced before the softmax operation, they alter the source-token locations receiving attention rather than merely rescaling the retrieved features afterward.

The two axial retrieval results are further coupled through the joint region direction representation. The joint embedding first produces feature-wise modulation parameters:
\begin{equation}
\left[
\gamma_{\mathrm{joint}};
\beta_{\mathrm{joint}}
\right]
=
W_Jc_{\mathrm{joint}},
\label{eq:joint_film_parameters}
\end{equation}
which are used to construct the Joint Axial Coupling pathway:
\begin{equation}
O_j^h
=
W_C^h
\left[
\left(
1+\gamma_{\mathrm{joint}}
\right)
\odot
\left(
O_v^h\odot O_h^h
\right)
+
\beta_{\mathrm{joint}}
\right].
\label{eq:joint_axial_coupling}
\end{equation}
This pathway allows the joint distribution to regulate how the vertically and horizontally retrieved evidence interacts, rather than treating the two axial pathways as independent contributions.

The morphology-routed axial pathways emphasize evidence associated with the dominant spinal curve but do not provide unrestricted access to the complete preoperative representation. To complement them with broader source information, DRICA includes a Global Context Retrieval pathway:
\begin{equation}
O_g^h
=
\operatorname{softmax}
\left(
\frac{Q_g^h(K_g^h)^{\top}}{\sqrt{d_h}}
\right)V_g^h.
\label{eq:global_retrieval}
\end{equation}
Unlike the axial pathways, Global Context Retrieval does not apply morphology-derived spatial biases and allows each current token to retrieve evidence from the complete preoperative token sequence. Its contribution is nevertheless adaptively controlled by the interval-dependent scale $s_g^h$ during route fusion.

After the four retrieval pathways have been computed, the joint morphology embedding and interval context determine the per-head fusion weights:
\begin{equation}
\left(
\alpha_v^h,
\alpha_h^h,
\alpha_j^h
\right)
=
\operatorname{softmax}
\left(
W_B^h
[c_{\mathrm{joint}};h_{r,t}]
\right).
\label{eq:branch_routing}
\end{equation}
The weights $\alpha_v^h$, $\alpha_h^h$, and $\alpha_j^h$ control the relative contributions of the vertical, horizontal, and joint axial pathways, while the Global Context Retrieval contribution is controlled by $s_g^h$.

The retrieved evidence is first fused across the different branches:
\begin{equation}
\widetilde{O}^h
=
s_{\mathrm{res}}^h
\left(
\alpha_v^hO_v^h
+
\alpha_h^hO_h^h
+
\alpha_j^hs_j^hO_j^h
+
s_g^hO_g^h
\right).
\label{eq:drica_branch_fusion}
\end{equation}
The fused evidence is then projected and written back to the current representation:
\begin{equation}
H_t'
=
H_t
+
W_O
\operatorname{Concat}_{h}
\left(
\widetilde{O}^h
\right).
\label{eq:drica_writeback}
\end{equation}
The interval-dependent scale $s_{\mathrm{res}}^h$ regulates the overall amount of source evidence written into the current stream. The output projection $W_O$ is initialized to zero, so each DRICA block initially behaves as an identity mapping and gradually learns to introduce preoperative evidence during training. The updated tokens $H_t'$ are then returned to the ViT$^{3}$ backbone for subsequent processing.

\subsection{Training and Inference}
\label{subsec:training}

ViT$^{3}$Flow is optimized using the finite-interval MeanFlow objective. For each sample in a mini-batch, two endpoint candidates are independently drawn using a logit-normal time sampler:
\begin{equation}
s_{i,j}
\overset{\mathrm{i.i.d.}}{\sim}
\mathcal{N}\left(\mu,\sigma^2\right).
\label{eq:logit_normal_sampling}
\end{equation}
Here, $j\in\{0,1\}$, $\mu=-0.4$, and $\sigma=1.0$. The sampled variables are transformed into time values as
\begin{equation}
u_{i,j}
=
\operatorname{sigmoid}\left(s_{i,j}\right).
\label{eq:logit_normal_transform}
\end{equation}
Thus, $u_{i,0},u_{i,1}\in(0,1)$. The two independently sampled values are then ordered to form a valid finite interval. The upper endpoint is defined as
\begin{equation}
\widetilde{t}_i
=
\max\left(u_{i,0},u_{i,1}\right),
\label{eq:ordered_upper_endpoint}
\end{equation}
and the lower endpoint is defined as
\begin{equation}
\widetilde{r}_i
=
\min\left(u_{i,0},u_{i,1}\right).
\label{eq:ordered_lower_endpoint}
\end{equation}
This ordering guarantees $\widetilde{t}_i\geq\widetilde{r}_i$. For each mini-batch of size $B$, a subset $\mathcal{S}$ containing $\lfloor\rho B\rfloor$ randomly selected samples is assigned equal endpoints:
\begin{equation}
(t_i,r_i)
=
\begin{cases}
(\widetilde{t}_i,\widetilde{t}_i), & i\in\mathcal{S},\\
(\widetilde{t}_i,\widetilde{r}_i), & i\notin\mathcal{S}.
\end{cases}
\label{eq:interval_mixture_sampling}
\end{equation}
We set $\rho=0.75$. Accordingly, approximately $75\%$ of the training samples satisfy $r=t$ and supervise the instantaneous flow-matching boundary, whereas the remaining $25\%$ use nonzero intervals $r<t$ to learn the interval-averaged MeanFlow velocity. All reported experiments use this logit-normal sampling configuration.

Along the probability path defined in Eq.~\eqref{eq:probability_path}, the total derivative of the predicted interval-averaged velocity is
\begin{equation}
\mathcal{D}_t u_{\theta}
=
\frac{\partial u_{\theta}}{\partial t}
+
J_z u_{\theta}v_t.
\label{eq:meanflow_total_derivative}
\end{equation}
The corresponding regression target is
\begin{equation}
u_{\mathrm{tgt}}
=
v_t-(t-r)\mathcal{D}_t u_{\theta}.
\label{eq:meanflow_target}
\end{equation}
Here, the derivative is taken along the probability path while the interval endpoint $r$, the preoperative image, and the SMEA-derived morphology conditions are held fixed. In practice, $\mathcal{D}_t u_{\theta}$ is evaluated using a Jacobian--vector product over $(z_t,t,r)$ with tangent $(v_t,1,0)$. For samples with $r=t$, the interval term vanishes and the target reduces to the conventional flow-matching target $u_{\mathrm{tgt}}=v_t$; for samples with $r<t$, the Jacobian--vector product provides the finite-interval correction required by the MeanFlow target.

For each sample $i$ in a mini-batch of size $B$, we define the
stop-gradient regression residual as
\begin{equation}
\varepsilon_i
=
u_{\theta}\left(z_t,t,r\mid\mathcal{C}\right)_i
-
\operatorname{sg}\left(u_{\mathrm{tgt},i}\right).
\label{eq:meanflow_residual}
\end{equation}
The regression error is then averaged over the image dimensions:
\begin{equation}
\Delta_i^2
=
\frac{1}{CHW}
\left\|
\varepsilon_i
\right\|_2^2.
\label{eq:meanflow_delta}
\end{equation}
Here, $\operatorname{sg}(\cdot)$ denotes the stop-gradient operation. The adaptive sample weight is defined as
\begin{equation}
w_i
=
\frac{1}
{\left(\Delta_i^2+\delta\right)^{1-\gamma}}.
\label{eq:meanflow_adaptive_weight}
\end{equation}
The corresponding mini-batch loss is
\begin{equation}
\ell_{\mathrm{adaptive}}
=
\frac{1}{B}
\sum_{i=1}^{B}
\operatorname{sg}(w_i)\Delta_i^2.
\label{eq:meanflow_adaptive_loss}
\end{equation}
We set $\gamma=0.5$ and $\delta=10^{-3}$. The detached weight $w_i$ rescales the contribution of each sample according to its current regression error without receiving gradients. The overall training objective is
\begin{equation}
\mathcal{L}_{\mathrm{MF}}
=
\mathbb{E}_{t,r,z_t,\mathcal{C}}
\left[
\ell_{\mathrm{adaptive}}
\right].
\label{eq:meanflow_objective}
\end{equation}
The expectation over $(t,r)$ follows the sampling procedure in Eqs.~\eqref{eq:logit_normal_sampling}--\eqref{eq:interval_mixture_sampling}.

During inference, the preoperative tokens $H_{\mathrm{pre}}$ and the
SMEA-derived morphology conditions are computed once and remain fixed
throughout the generation trajectory. Starting from
$z_{t_0}\sim\mathcal{N}(0,I)$ with $t_0=1$, we use a decreasing schedule
$1=t_0>t_1>\cdots>t_N=0$ and define the interval length as
$\Delta t_n=t_n-t_{n+1}$. The state is updated as
\begin{equation}
z_{t_{n+1}}
=
z_{t_n}
-
\Delta t_n\,
u_{\theta}
\left(
z_{t_n},t_n,t_{n+1}\mid\mathcal{C}
\right).
\label{eq:inference_update}
\end{equation}
The update is applied for $n=0,\ldots,N-1$. Although the morphology
conditions remain fixed for a given case, their effect on source
retrieval varies across intervals through the interval-dependent
controls in DRICA. In the principal single-NFE setting, the schedule
contains only $(t_0,t_1)=(1,0)$, and the postoperative state is obtained
from one evaluation of
$u_{\theta}(z_1,1,0\mid\mathcal{C})$. Optional few-step inference
applies the same update recursively over a partitioned schedule. The
final postoperative prediction $\hat{x}_{\mathrm{post}}$ is obtained by
denormalizing $z_0$ and clipping it to the valid image range.

%% file: sections/4_experiments.tex
\section{Results}
\label{sec:experiments}

\subsection{Dataset and implementation details}

\subsubsection{Datasets}
\label{subsubsec:datasets}

We introduce ScoliSurg, a paired imaging dataset for scoliosis corrective surgery comprising 632 matched preoperative and postoperative standing whole-spine radiograph pairs. The data were retrospectively collected from two collaborating hospitals, together with demographic information and clinical diagnostic records. Patients were included if they had a clinically diagnosed spinal deformity, underwent corrective surgery, and had complete preoperative and postoperative imaging records. Patients with psychological disorders, trauma affecting posture or mobility, severe dermatological conditions interfering with imaging, or known oncological conditions were excluded. All radiograph pairs were screened for image quality, correct patient-level correspondence, comparable anatomical coverage, and reliable postoperative assessment by two senior spine surgeons, each with more than 10 years of clinical experience. The study protocol was approved by the institutional ethics review board, and written informed consent was obtained from all participants or their legal guardians.

\subsubsection{Implementation details}
We conduct all experiments on ScoliSurg using a 7:2:1 train/validation/test split. All radiographs are resized to $480\times240$ ($H\times W$). All methods are implemented in PyTorch and optimized using the AdamW optimizer~\cite{loshchilov2019adamw} with a learning rate of $1\times10^{-4}$. ViT$^{3}$Flow is optimized using the adaptive finite-interval MeanFlow loss defined in Eq.~\eqref{eq:meanflow_objective}. Models are trained for $5\times10^{6}$ steps with a batch size of 4. For the ViT$^{3}$ token mixer, the inner model is initialized from the learned outer parameters and independently adapted for each input sample. Following the original ViT$^{3}$ setting, we perform a single full-batch inner update using all token key--value pairs, with a dot-product reconstruction objective and an inner-loop learning rate of 0.25. The adapted parameters are reset to the shared initialization before processing the next sample. We additionally apply 2D axial rotary positional embeddings to preserve spatial token relationships. All experiments are performed on a cloud platform equipped with four NVIDIA RTX A6000 GPUs.

\subsubsection{Evaluation metrics}
\label{subsubsec:evaluation_metrics}

We evaluate image quality using peak signal-to-noise ratio (PSNR), structural similarity index measure (SSIM), and learned perceptual image patch similarity (LPIPS)~\cite{zhang2018unreasonable}. Clinical accuracy is assessed using the mean absolute error (MAE) of the postoperative Cobb angle~\cite{nowitzke2008improving} and correction rate (CR) error. For each test case, the Cobb angles on the preoperative radiograph, synthesized postoperative radiograph, and corresponding ground-truth postoperative radiograph are independently measured by two senior spine surgeons following the same measurement protocol. The final Cobb angle for each radiograph is obtained by averaging the measurements from the two surgeons. The postoperative Cobb angle MAE is then calculated between the averaged measurements of the synthesized and ground-truth postoperative radiographs. CR is defined as
\begin{equation}
\mathrm{CR}
=
\frac{\theta_{\mathrm{pre}}-\theta_{\mathrm{post}}}
{\theta_{\mathrm{pre}}}
\times 100\%,
\label{eq:correction_rate}
\end{equation}
where $\theta_{\mathrm{pre}}$ and $\theta_{\mathrm{post}}$ denote the averaged manually measured preoperative and postoperative Cobb angles, respectively~\cite{langensiepen2013cobb}. The CR error is calculated as the absolute difference between the CR values derived from the synthesized and ground-truth postoperative radiographs. For each test case, the absolute postoperative Cobb-angle error and CR error are calculated independently, and the reported values are summarized as mean $\pm$ standard deviation across all individual cases in the test set.

\subsection{Comparison with State-of-the-Art Methods}
\label{subsec:sota_comparison}

\subsubsection{Quantitative Comparison}
\label{subsec:quantitative_comparison}

Postoperative radiograph synthesis requires both faithful reconstruction of radiographic appearance and accurate representation of the surgically corrected spinal geometry. Table~\ref{tab:main_comparison} compares ViT$^{3}$Flow with multi-NFE diffusion- and flow-based models as well as single-NFE image-translation methods. The multi-NFE baselines exhibit complementary but fragmented strengths. Diffusion achieves the highest baseline PSNR of 25.81~dB, $I^2$SB obtains the lowest baseline LPIPS of 0.1958, and FM provides the strongest postoperative morphology measurements, with a Cobb angle MAE of $1.52\pm3.57^\circ$ and a CR error of $2.98\pm4.76$~pp. No individual baseline, however, performs consistently well across both image-quality and morphology-related metrics. The single-NFE image-translation methods similarly provide efficient generation but show less consistent preservation of postoperative structure. RegGAN, for example, achieves a relatively low Cobb angle MAE of $1.62\pm2.58^\circ$, but its LPIPS remains high at 0.2649, indicating that geometric accuracy does not necessarily translate into perceptually faithful reconstruction.

ViT$^{3}$Flow achieves a more consistent balance between these two requirements while retaining single-NFE inference. It obtains the best result across all five metrics, reaching a PSNR of 26.79~dB and an LPIPS of 0.1828 while reducing the Cobb angle MAE and CR error to $1.07\pm2.30^\circ$ and $2.23\pm4.71$~pp, respectively. Compared with FM, the strongest baseline for postoperative morphology, ViT$^{3}$Flow reduces Cobb angle MAE by 29.6\% and CR error by 25.2\%, while simultaneously improving PSNR from 25.42 to 26.79~dB and LPIPS from 0.2524 to 0.1828. It also surpasses the strongest baseline values separately obtained by Diffusion and $I^2$SB for PSNR and LPIPS. These results show that single-NFE generation does not require sacrificing either radiographic fidelity or postoperative morphological accuracy, and that ViT$^{3}$Flow provides a unified improvement over methods whose advantages are limited to individual evaluation dimensions.

\begin{table*}[t]
\centering
\caption{Quantitative comparison with image-translation, diffusion-based, and flow-based methods on ScoliSurg. NFE denotes the number of network function evaluations during inference. Higher PSNR and SSIM are better, while lower LPIPS, Cobb MAE, and CR error are better.}
\label{tab:main_comparison}
\setlength{\tabcolsep}{5pt}
\renewcommand{\arraystretch}{1.08}

{\fontsize{10pt}{12pt}\selectfont
\begin{tabular}{@{}lcccccc@{}}
\toprule
Method & NFE & PSNR $\uparrow$ & SSIM $\uparrow$ & LPIPS $\downarrow$ &
Cobb MAE ($^\circ$) $\downarrow$ & CR Error (pp) $\downarrow$ \\
\midrule
\multicolumn{7}{c}{\textit{Multi-NFE Methods}} \\
\midrule
Palette~\cite{saharia2022palette}
& 1000 & 22.18 & 0.7845 & 0.2189
& 9.38$\pm$7.61 & 16.81$\pm$8.01 \\

$I^2$SB~\cite{liu2023i2sb}
& 100 & 25.54 & 0.7926 & 0.1958
& 8.83$\pm$2.49 & 13.13$\pm$4.25 \\

Diffusion~\cite{ho2020denoisingdiffusionprobabilisticmodels}
& 1000 & 25.81 & 0.8237 & 0.2416
& 2.85$\pm$3.95 & 6.43$\pm$8.55 \\

MOTFM~\cite{yazdani2025flow}
& 10 & 24.40 & 0.8031 & 0.2223
& 4.79$\pm$3.25 & 6.33$\pm$5.08 \\

FM~\cite{lipman2023flowmatchinggenerativemodeling}
& 100 & 25.42 & 0.8319 & 0.2524
& 1.52$\pm$3.57 & 2.98$\pm$4.76 \\

\midrule
\multicolumn{7}{c}{\textit{Single-NFE Methods}} \\
\midrule
Pix2Pix~\cite{isola2018imagetoimagetranslationconditionaladversarial}
& 1 & 24.76 & 0.7826 & 0.2635
& 5.70$\pm$4.86 & 12.34$\pm$12.02 \\

CycleGAN~\cite{zhu2020unpairedimagetoimagetranslationusing}
& 1 & 24.50 & 0.7833 & 0.2720
& 3.45$\pm$4.82 & 7.15$\pm$9.74 \\

CUT~\cite{park2020contrastive}
& 1 & 23.34 & 0.7568 & 0.3078
& 11.26$\pm$18.03 & 20.48$\pm$30.12 \\

RegGAN~\cite{kong2021reggan}
& 1 & 25.62 & 0.8015 & 0.2649
& 1.62$\pm$2.58 & 3.14$\pm$5.27 \\

ResViT~\cite{dalmaz2022resvit}
& 1 & 24.45 & 0.7790 & 0.2838
& 4.90$\pm$6.05 & 10.21$\pm$13.42 \\

\midrule
\rowcolor{gray!12}
\textbf{Ours}
& 1
& \textbf{26.79}
& \textbf{0.8641}
& \textbf{0.1828}
& \textbf{1.07$\pm$2.30}
& \textbf{2.23$\pm$4.71} \\
\bottomrule
\end{tabular}
}
\end{table*}

\subsubsection{Efficiency Comparison}
\label{subsec:efficiency_comparison}

\begin{figure}[pos=!t]
\centering
\includegraphics[width=\columnwidth]{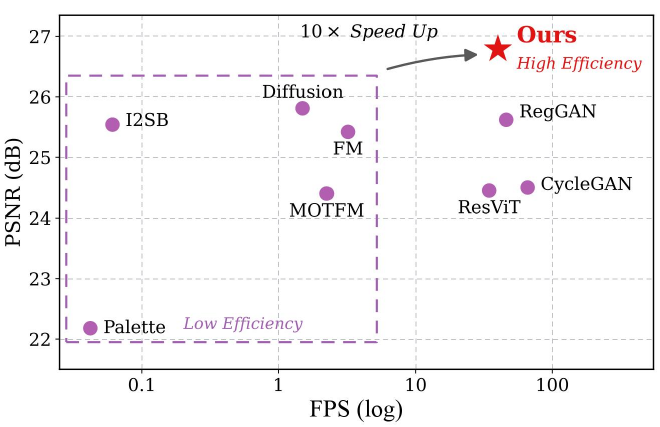}
\vspace{-2mm}
\caption{Quality efficiency comparison on ScoliSurg. Reconstruction quality is measured by PSNR, and inference throughput is reported in frames per second (FPS) on a logarithmic scale. Multi-NFE diffusion- and flow-based methods are concentrated in the low-throughput region, whereas single-NFE image-translation methods achieve higher throughput but lower reconstruction quality. ViT$^{3}$Flow reaches the highest PSNR while remaining in the high-efficiency regime.}
\vspace{-3mm}
\label{fig:efficiency}
\end{figure}

As shown in Fig.~\ref{fig:efficiency}, multi-NFE diffusion- and flow-based methods are concentrated in the low-throughput region because they repeatedly evaluate the generative network along the sampling trajectory. Although Diffusion reaches the strongest baseline PSNR of 25.81~dB, its inference speed remains substantially lower than that of the single-NFE methods. In contrast, image-translation models operate at tens of frames per second but generally achieve lower reconstruction quality; RegGAN obtains the highest PSNR among these methods at 25.62~dB, while ResViT and CycleGAN reach 24.45 and 24.50~dB, respectively.

ViT$^{3}$Flow achieves a more favorable quality-efficiency balance, reaching the highest PSNR of 26.79~dB while retaining single-NFE inference and high throughput. It exceeds RegGAN, the strongest single-NFE baseline in PSNR, by 1.17~dB and operates approximately one order of magnitude faster than representative multi-NFE diffusion- and flow-based methods. These results indicate that ViT$^{3}$Flow improves reconstruction quality without relying on a long iterative sampling process.

\subsubsection{Visual Comparison}
\label{subsec:visual_comparison}

\begin{figure}[pos=!t]
\centering
\includegraphics[width=\columnwidth]{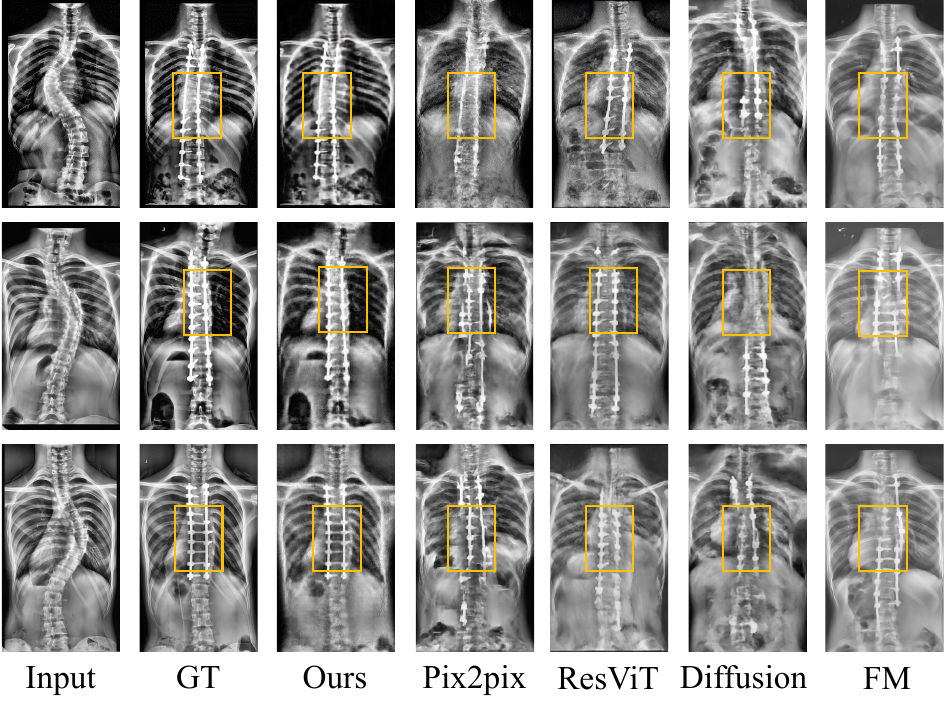}
\vspace{-2mm}
\caption{Qualitative comparison of postoperative scoliosis radiograph synthesis on three representative cases. From left to right: preoperative input, postoperative ground truth (GT), ViT$^{3}$Flow prediction, Pix2Pix, ResViT, diffusion-based model, and flow-matching model (FM). The yellow boxes highlight the principal correction and fixation regions. ViT$^{3}$Flow more closely reproduces the corrected spinal trajectory and the continuity of postoperative fixation structures while preserving the surrounding anatomical appearance.}
\vspace{-3mm}
\label{fig:visual_comparison}
\end{figure}

Figure~\ref{fig:visual_comparison} presents three representative cases with different preoperative curvature patterns. In the corresponding ground-truth postoperative radiographs, the spinal axis is substantially corrected, and continuous fixation structures are visible within the highlighted regions.

Pix2Pix captures the coarse postoperative appearance but produces blurred vertebral boundaries, incomplete fixation structures, and residual curvature in several cases. ResViT better preserves the overall body contour but introduces local inconsistencies in the reconstructed spinal axis and fixation pattern. The diffusion-based model recovers part of the corrected configuration, although the highlighted regions contain discontinuous or poorly localized structural details. FM produces more apparent fixation-related features, but the surrounding vertebral and soft-tissue structures are frequently over-smoothed, and the reconstructed alignment does not consistently match the ground truth.

In contrast, ViT$^{3}$Flow more closely reproduces the corrected spinal trajectory and maintains a more continuous fixation pattern across the three cases. The highlighted regions also show closer correspondence with the ground-truth vertebral configuration while preserving the surrounding anatomical appearance. These observations are consistent with the lower Cobb angle MAE and CR error reported in Table~\ref{tab:main_comparison}.

\subsection{Ablation Study}
\label{subsec:ablation}

\subsubsection{Contribution of Core Components}
\label{subsec:component_ablation}

We evaluate the individual and combined contributions of the finite-interval MeanFlow objective, the ViT$^{3}$ token mixer, and DRICA under the same network scale, training schedule, and inference setting. The Base model adopts conventional flow matching and standard self-attention. M1 replaces the training objective with MeanFlow, M2 further introduces the ViT$^{3}$ token mixer without DRICA or source-token cross-attention, and M3 combines MeanFlow with DRICA while retaining the conventional token mixer. The complete model integrates all three components.

\begin{table*}[t]
\centering
\caption{Contribution analysis of the core components in ViT$^{3}$Flow on ScoliSurg. Higher PSNR and SSIM are better, while lower LPIPS, Cobb MAE, and CR error are better.}
\label{tab:main_ablation}
\setlength{\tabcolsep}{5pt}
\renewcommand{\arraystretch}{1.08}

{\fontsize{10.5pt}{12.5pt}\selectfont
\begin{tabular}{@{}lccc|ccccc@{}}
\toprule
Variant & MeanFlow & ViT$^{3}$ & DRICA & PSNR $\uparrow$ & SSIM $\uparrow$ & LPIPS $\downarrow$ & Cobb MAE ($^\circ$) $\downarrow$ & CR Error (pp) $\downarrow$ \\
\midrule
Base & & & & 25.42 & 0.8319 & 0.2524 & 1.52$\pm$3.57 & 2.98$\pm$4.76 \\
M1 & \checkmark & & & 25.62 & 0.8495 & 0.2311 & 1.56$\pm$2.47 & 2.69$\pm$3.50 \\
M2 & \checkmark & \checkmark & & 26.20 & 0.8596 & 0.1961 & 1.67$\pm$3.34 & 2.87$\pm$2.83 \\
M3 & \checkmark & & \checkmark & 25.59 & 0.8468 & 0.2097 & 1.11$\pm$2.44 & 2.84$\pm$3.01 \\
\midrule
\rowcolor{gray!12}
Ours & \checkmark & \checkmark & \checkmark & \textbf{26.79} & \textbf{0.8641} & \textbf{0.1828} & \textbf{1.07$\pm$2.30} & \textbf{2.23$\pm$4.71} \\
\bottomrule
\end{tabular}
}
\end{table*}

Table~\ref{tab:main_ablation} shows that the three components contribute to complementary aspects of postoperative radiograph synthesis. Replacing conventional flow matching with the finite-interval MeanFlow objective in M1 increases PSNR from 25.42 to 25.62~dB, improves SSIM from 0.8319 to 0.8495, reduces LPIPS from 0.2524 to 0.2311, and lowers CR error from 2.98 to 2.69~pp. These consistent improvements demonstrate the effectiveness of finite-interval supervision for learning the postoperative transformation.

The contribution of the ViT$^{3}$ token mixer is particularly evident in the image-quality metrics. Relative to M1, M2 increases PSNR from 25.62 to 26.20~dB, improves SSIM from 0.8495 to 0.8596, and reduces LPIPS from 0.2311 to 0.1961. These gains show that test-time-training token dynamics strengthens the representation of the evolving postoperative state and improves radiographic reconstruction quality.

DRICA provides a complementary improvement in morphology related accuracy. Compared with M1, M3 reduces Cobb angle MAE from $1.56\pm2.47^\circ$ to $1.11\pm2.44^\circ$ and improves LPIPS from 0.2311 to 0.2097. This pattern highlights the value of structured retrieval from the preoperative token stream for maintaining patient-specific anatomical correspondence during postoperative synthesis.

Integrating MeanFlow, ViT$^{3}$, and DRICA produces the best performance across all five metrics. The complete model reaches a PSNR of 26.79~dB, an SSIM of 0.8641, and an LPIPS of 0.1828, while further reducing Cobb angle MAE and CR error to $1.07\pm2.30^\circ$ and $2.23\pm4.71$~pp, respectively. These results demonstrate a complementary relationship among the three components: MeanFlow supports efficient finite-interval generation, ViT$^{3}$ enhances test-time-training token dynamics, and DRICA introduces spatially organized preoperative evidence into the evolving postoperative representation.

\subsubsection{Analysis of DRICA Routing Mechanisms}
\label{subsec:drica_ablation}

We further analyze the contributions of Global Context Retrieval, axial retrieval, SMEA-guided morphology routing, interval-dependent modulation, and joint axial coupling within DRICA. All variants use the same MeanFlow objective and ViT$^{3}$ backbone. R1 retains the DRICA insertion structure but uses only Global Context Retrieval from the evolving current tokens to the complete preoperative token sequence, while R2 further introduces axial decomposition. R3--R5 retain both the global and axial retrieval pathways but individually remove the morphology guidance provided by the Spinal Morphology Extraction Agent (SMEA), interval-dependent modulation, and joint axial coupling from the complete configuration in R6, respectively. Specifically, SMEA extracts the dominant-curve region and direction distributions that guide the spatial routing of preoperative evidence in DRICA. R1 also differs from M2 in Table~\ref{tab:main_ablation}: M2 does not use SMEA or source-token cross-attention and instead incorporates only a pooled preoperative condition through adaptive layer normalization (AdaLN).

\begin{table*}[t]
\centering
\small
\caption{Analysis of the routing mechanisms in Diagnosis Routed Interval Cross Attention (DRICA). Higher PSNR and SSIM are better, while lower LPIPS, Cobb MAE, and CR error are better.}
\label{tab:drica_ablation}
\setlength{\tabcolsep}{4pt}
\renewcommand{\arraystretch}{1.08}
\resizebox{\textwidth}{!}{%
\begin{tabular}{@{}lccccc|ccccc@{}}
\toprule
Variant & Global Ctx. & Axial & Morph. & Interval & Joint & PSNR $\uparrow$ & SSIM $\uparrow$ & LPIPS $\downarrow$ & Cobb MAE ($^\circ$) $\downarrow$ & CR Error (pp) $\downarrow$ \\
\midrule
R1 & \checkmark & & & & & 25.23 & 0.8394 & 0.2102 & 2.58$\pm$2.81 & 3.88$\pm$2.75 \\
R2 & \checkmark & \checkmark & & & & 25.41 & 0.8355 & 0.2035 & 2.17$\pm$3.29 & 3.77$\pm$2.82 \\
R3 & \checkmark & \checkmark & & \checkmark & \checkmark & 26.46 & 0.8461 & 0.1975 & 2.01$\pm$3.73 & 2.55$\pm$2.58 \\
R4 & \checkmark & \checkmark & \checkmark & & \checkmark & 26.42 & 0.8544 & 0.1969 & 2.03$\pm$3.02 & 3.56$\pm$3.06 \\
R5 & \checkmark & \checkmark & \checkmark & \checkmark & & 26.28 & 0.8505 & 0.1893 & 1.46$\pm$4.08 & 2.47$\pm$2.47 \\
\midrule
\rowcolor{gray!12}
R6 & \checkmark & \checkmark & \checkmark & \checkmark & \checkmark & \textbf{26.79} & \textbf{0.8641} & \textbf{0.1828} & \textbf{1.07$\pm$2.30} & \textbf{2.23$\pm$4.71} \\
\bottomrule
\end{tabular}%
}
\end{table*}

Comparisons across R1-R6 demonstrate the complementary roles of the DRICA routing mechanisms. Relative to the global-only retrieval in R1, introducing axial decomposition in R2 increases PSNR from 25.23 to 25.41~dB, reduces LPIPS from 0.2102 to 0.2035, and lowers Cobb angle MAE from $2.58\pm2.81^\circ$ to $2.17\pm3.29^\circ$, showing the benefit of organizing preoperative evidence retrieval along the vertical and horizontal axes of the whole-spine image. The SMEA-derived morphology conditions provide further patient-specific guidance: compared with R3, which excludes the region and direction distributions, the complete configuration R6 improves PSNR from 26.46 to 26.79~dB, reduces LPIPS from 0.1975 to 0.1828, and lowers Cobb angle MAE from $2.01\pm3.73^\circ$ to $1.07\pm2.30^\circ$.

Interval-dependent modulation and joint axial coupling contribute additional complementary improvements. Compared with R4, R6 improves all five metrics and reduces the mean CR error from $3.56$ to $2.23$~pp, supporting the use of MeanFlow interval context to regulate route scales and source-evidence write-back. Compared with R5, introducing joint axial coupling improves PSNR from 26.28 to 26.79~dB, SSIM from 0.8505 to 0.8641, and LPIPS from 0.1893 to 0.1828, while reducing Cobb angle MAE from $1.46\pm4.08^\circ$ to $1.07\pm2.30^\circ$. Overall, R6 combines global access, axial organization, SMEA-guided morphology routing, interval modulation, and cross-axis interaction, achieving the best performance across all image-quality and morphology-related metrics.

\subsection{Clinical Correlation and Agreement Analysis}
\label{subsec:clinical_agreement}

To assess whether the synthesized radiographs preserve clinically relevant postoperative spinal curvature, we compare the Cobb angles measured from the synthesized images with those measured from the corresponding ground-truth postoperative radiographs. Pearson and Spearman correlation analyses are used to evaluate the association between the paired measurements, while Bland--Altman analysis quantifies their mean difference and 95\% limits of agreement.

\begin{figure*}[pos=t]
\centering
\includegraphics[width=\textwidth]{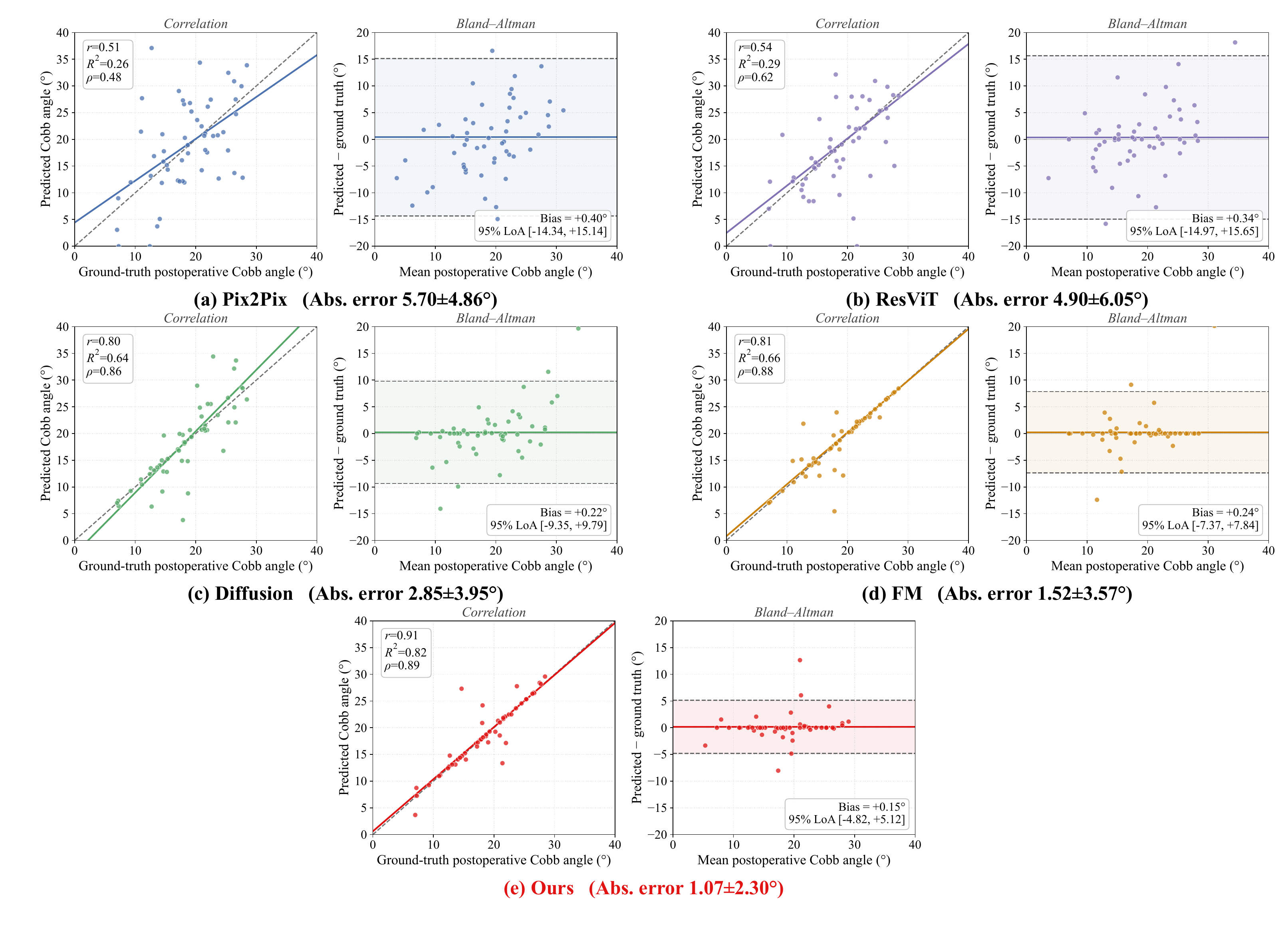}
\vspace{-2mm}
\caption{Clinical correlation and agreement analysis of postoperative Cobb angle measurements for Pix2Pix, ResViT, Diffusion, FM, and ViT$^{3}$Flow. For each method, the correlation plot compares the predicted and ground-truth postoperative Cobb angles. The dashed diagonal line denotes the identity line, while the solid colored line denotes the fitted regression line. Pearson correlation coefficient ($r$), coefficient of determination ($R^2$), and Spearman correlation coefficient ($\rho$) are reported in each plot, and the corresponding absolute angular error is shown below each method. The adjacent Bland--Altman plot presents the difference between the predicted and ground-truth Cobb angles against their paired mean. The solid horizontal line indicates the mean difference, and the dashed horizontal lines indicate the 95\% limits of agreement.}
\vspace{-3mm}
\label{fig:cobb_agreement}
\end{figure*}

As shown in Fig.~\ref{fig:cobb_agreement}, ViT$^{3}$Flow demonstrates the strongest association with the ground-truth postoperative Cobb angle measurements, achieving $r=0.91$, $R^2=0.82$, and $\rho=0.89$, together with the lowest absolute angular error of $1.07\pm2.30^\circ$. These results indicate that the Cobb angles measured from the synthesized radiographs closely follow the variation observed in the corresponding postoperative measurements.

The Bland--Altman analysis further shows that ViT$^{3}$Flow has a mean measurement difference of $0.15^\circ$, with 95\% limits of agreement ranging from $-4.82^\circ$ to $5.12^\circ$. These limits are narrower than those of the competing methods. For example, FM yields limits of agreement from $-7.37^\circ$ to $7.84^\circ$ and a lower Pearson correlation of $r=0.81$. Overall, ViT$^{3}$Flow achieves stronger correlation and closer agreement with ground-truth Cobb angle measurements, supporting its ability to preserve clinically relevant postoperative spinal morphology.

%% file: sections/5_discussion.tex
\section{Discussion}
\label{sec:discussion}

This study presents ViT$^{3}$Flow for efficient postoperative scoliosis radiograph synthesis. The proposed framework integrates finite-interval MeanFlow generation, test-time-training token dynamics, and diagnosis-routed retrieval of preoperative evidence. Evaluations based on conventional image-quality metrics and postoperative Cobb angle measurements demonstrate that ViT$^{3}$Flow improves radiographic fidelity and morphological accuracy under single-NFE inference. Ablation studies on the main framework components and the internal routing mechanisms of DRICA further show that finite-interval generation, test-time-training token dynamics, and diagnosis-routed retrieval provide complementary improvements.

Two factors are particularly important for postoperative radiograph synthesis. The first is the efficient modeling of large corrective changes. Conventional diffusion- and bridge-based methods approximate the generative trajectory through multiple local updates, resulting in greater inference cost and potentially reduced structural accuracy when the number of sampling steps is substantially decreased. ViT$^{3}$Flow instead predicts an interval-averaged velocity over a nonzero transport interval, allowing a single network evaluation to represent a larger portion of the transformation from noise to the postoperative state. Comparisons with both multi-NFE and single-NFE baselines show that this formulation achieves a favorable balance between synthesis quality and inference efficiency.

The second factor is the selective use of preoperative anatomy. Surgical correction substantially alters spinal alignment, whereas vertebral appearance and surrounding anatomical structures should remain consistent with the individual patient. Global conditioning alone may not adequately distinguish regions requiring substantial correction from structures whose anatomical identity should be retained. In ViT$^{3}$Flow, the ViT$^{3}$ mixer adapts token interactions to the current case, while DRICA retrieves spatially corresponding evidence from a separate preoperative token stream according to the dominant-curve distributions and transport interval. The ablation results support these complementary roles: test-time-training token dynamics mainly improves radiographic reconstruction, whereas source-evidence retrieval provides additional gains in postoperative morphological accuracy. Their integration yields the strongest overall performance across image quality and clinically relevant geometric measurements.

From a clinical perspective, ViT$^{3}$Flow provides an image-level estimate of postoperative spinal morphology from the available preoperative radiograph. By representing the spatial distribution of correction across the complete radiograph, the synthesized result may complement conventional scalar outcome estimates and support postoperative alignment assessment, correction analysis, and outcome evaluation. Incorporating operative information, such as the planned correction strategy and target alignment, may further improve the specificity and controllability of future models.

The present study has several limitations. ScoliSurg was retrospectively collected in a single clinical setting, and external validation on data from additional institutions is required to assess generalizability. Moreover, the current evaluation focuses on two-dimensional radiographic morphology and selected image-quality and geometric measurements. Future work will expand the dataset across clinical centers, incorporate surgical conditions, and investigate the utility of synthesized radiographs in downstream quantitative and clinical assessment tasks.

%% file: sections/6_conclusion.tex
\section{Conclusion}
\label{sec:conclusion}

In this paper, we introduce ScoliSurg, a paired dataset for postoperative scoliosis radiograph synthesis, and propose ViT$^{3}$Flow, a single-NFE framework for generating postoperative radiographs from preoperative images. ViT$^{3}$Flow formulates the task as patient-conditioned finite-interval transport and combines MeanFlow with sample-adaptive token computation and structured retrieval of preoperative anatomy. The Spinal Morphology Extraction Agent extracts distributions of dominant-curve region and direction from the preoperative radiograph, which guide Diagnosis Routed Interval Cross Attention (DRICA) in retrieving spatially corresponding source evidence, while the ViT$^{3}$ token mixer adapts its computation to the morphology of each case.

Experiments on ScoliSurg show that ViT$^{3}$Flow achieves strong radiographic fidelity and postoperative morphological accuracy with a single network evaluation, outperforming representative direct-translation, diffusion-, bridge-, and flow-based baselines. Ablation studies further demonstrate the distinct and complementary roles of finite-interval generation, sample-specific token adaptation, and diagnosis-routed source retrieval. These results establish the feasibility of efficient postoperative radiograph synthesis. Future work will extend the evaluation to multi-center cohorts, incorporate planned surgical variables to improve generation specificity and controllability, and assess the utility of synthesized radiographs for quantitative postoperative analysis and clinical decision support.